\documentclass{article}

\PassOptionsToPackage{numbers, compress}{natbib}

\usepackage[preprint]{neurips_2025}

\usepackage[utf8]{inputenc} 
\usepackage[T1]{fontenc}    

\usepackage{url}            
\usepackage{booktabs}       
\usepackage{amsfonts}       
\usepackage{nicefrac}       
\usepackage{microtype}      
\usepackage[table,xcdraw]{xcolor}

\usepackage{wrapfig}
\usepackage{algorithm}
\usepackage{algorithmic}
\usepackage{amsmath,amsfonts}
\usepackage{adjustbox} 
\usepackage{algorithmic}
\usepackage{array}
\usepackage[caption=false,font=normalsize,labelfont=sf,textfont=sf]{subfig}
\usepackage{textcomp}
\usepackage{url}
\usepackage{verbatim}
\usepackage{graphicx}
\usepackage{textcomp} 
\usepackage{amsthm,amsmath,amssymb}
\usepackage{mathrsfs}
\usepackage{booktabs}
\usepackage{enumerate}
\usepackage{bbding}
\usepackage{multirow} 
\usepackage{ulem}
\usepackage{utfsym}
\usepackage{times}
\usepackage{latexsym}
\usepackage{inconsolata}
\usepackage{mathrsfs}
\usepackage{adjustbox}
\usepackage{multirow}
\useunder{\uline}{\ul}{}

\usepackage[pdftex]{hyperref}

\title{HSENet: Hybrid Spatial Encoding Network for 3D Medical Vision-Language Understanding}

\author{
    Yanzhao Shi\textsuperscript{1},
    Xiaodan Zhang\textsuperscript{2},
    Junzhong Ji\textsuperscript{2},
    \textbf{Haoning Jiang}\textsuperscript{\textbf{1}},\\
    \textbf{Chengxin Zheng}\textsuperscript{\textbf{2}},
    \textbf{Yinong Wang}\textsuperscript{\textbf{1}},
    \textbf{Liangqiong Qu}\textsuperscript{\textbf{1}}\\
\textsuperscript{1} The University of Hong Kong, Hong Kong\\
\textsuperscript{2} Beijing University of Technology, Beijing, China\\
    zhangxiaodan@bjut.edu.cn, liangqqu@hku.hk
}

\begin{document}

\maketitle

\begin{abstract}
Automated 3D CT diagnosis empowers clinicians to make timely, evidence-based decisions by enhancing diagnostic accuracy and workflow efficiency. While multimodal large language models (MLLMs) exhibit promising performance in visual-language understanding, existing methods mainly focus on 2D medical images, which fundamentally limits their ability to capture complex 3D anatomical structures. This limitation often leads to misinterpretation of subtle pathologies and causes diagnostic hallucinations. In this paper, we present Hybrid Spatial Encoding Network (HSENet), a framework that exploits enriched 3D medical visual cues by effective visual perception and projection for accurate and robust vision-language understanding. Specifically, HSENet employs dual-3D vision encoders to perceive both global volumetric contexts and fine-grained anatomical details, which are pre-trained by dual-stage alignment with diagnostic reports. Furthermore, we propose Spatial Packer, an efficient multimodal projector that condenses high-resolution 3D spatial regions into a compact set of informative visual tokens via centroid-based compression. By assigning spatial packers with dual-3D vision encoders, HSENet can seamlessly perceive and transfer hybrid visual representations to LLM's semantic space, facilitating accurate diagnostic text generation. Experimental results demonstrate that our method achieves state-of-the-art performance in 3D language-visual retrieval (39.85\% of R@100, +5.96\% gain), 3D medical report generation (24.01\% of BLEU-4, +8.01\% gain), and 3D visual question answering (73.60\% of Major Class Accuracy, +1.99\% gain), confirming its effectiveness. 
Our code is available at~\href{https://github.com/YanzhaoShi/HSENet}{\textit{https://github.com/YanzhaoShi/HSENet}}.

\end{abstract}

\section{Introduction}
3D computed tomography (CT) has revolutionized medical diagnostics by providing high-resolution visualization of anatomical structures. Nonetheless, interpreting 3D CT images is labor-intensive for radiologists, which relies heavily on intricate psychophysiological and cognitive processes that are prone to perceptual errors~\cite{Bruno2015UnderstandingAC}.
The application of computer-aided diagnostic models offers considerable promise in assisting radiologists for efficient and accurate clinical decision-making.

Recently, multi-modal large language models (MLLMs) have emerged as a powerful tool in medical image analysis, including diagnostic tasks such as medical report generation (MRG) and visual question answering (VQA). 
Current works mainly focus on 2D medical imaging, such as X-ray~\cite{jing2018automatic,wang2022cross,li2023dynamic,tanno2025collaboration}, which offers planar projections valuable for screening thoracic conditions and skeletal disorders. 
However, 2D imaging inherently fails to capture volumetric details of complex anatomical relationships, restricting the ability of MLLMs to interpret spatial patterns in lesions.
This restriction hinders their clinical utility of models in scenarios requiring volumetric analysis, such as tumor infiltration assessment or vascular anomaly detection.
To address this challenge, early studies shift toward 3D CT imaging, employing slice-by-slice analysis~\cite{Li2024TowardsAH,Zhang2025MEPNet} or in chunks of small stacks of 2D slices~\cite{Huang2020PENet}, yet these methods still struggle to capture spatial continuity along the depth (z-axis) dimension.
In contrast, RadFM~\cite{Wu23RadFM} and M3D~\cite{Bai24M3D} leverage 3D Vision Transformers (ViTs) to train foundation MLLMs, utilizing a large volume of 3D medical samples to enhance the model adaptability across various tasks.
To further reduce diagnostic hallucinations and improve clinical performance, these foundation models are integrated with specialized visual pretraining strategies~\cite{Xin2025Med3DVLMAE,Lai2025Bridged,Ni2024MG3D} and visual encoding pipelines~\cite{Shi24Med2E3,Chen2024Dia,Chen20243DCTGPT}.
Nevertheless, existing methods still encounter challenges in understanding spatial details of 3D anatomical structures due to several key issues:

\textbf{Limited visual perception.} 
CLIP-style vision encoders~\cite{Bai24M3D, Hamamci2024DevelopingGF, Zhang2024BiomedCLIPAM, Xin2025Med3DVLMAE} are commonly utilized to extract discriminative visual features aligned with expert reports.
However, unlike natural image-report datasets (e.g., 400M pairs~\cite{Radford2021CLIP}), the scarcity of 3D volume-report pairs (roughly 0.05M~\cite{Hamamci2024DevelopingGF}) highly constrains feature space convergence.
As a result, subtle but clinically critical pathological details may be obscured by irrelevant information, leading to suboptimal visual interpretation.

\textbf{Compromised semantic projection.} 
While multi-modal projectors aim to bridge vision and language by mapping 3D visual representations into LLM semantic spaces, current approaches (e.g., spatial pooling~\cite{Bai24M3D} and Q-former~\cite{Li2023BLIP2,Chen2023MedBLIP}) struggle to preserve spatial and geometric details inherent in 3D anatomical structures.
This limitation undermines the ability of LLMs to reason structural dependencies and pathological conditions, leading to unreliable outputs with fundamental errors.

In this paper, we propose Hybrid Spatial Encoding Network (HSENet), a novel framework that exploits enriched 3D medical visual cues with effective visual perception and projection for robust vision-language understanding.
Specifically, to perceive spatial contexts from 3D volumetric space, we introduce a dual-stage 3D vision-language pretraining paradigm that trains dual-3D vision encoders: 
A 3D Vision Encoder learns global volumetric representations aligned with corresponding reports, while a 2D-Enhanced 3D Vision Encoder (2E3 Vision Encoder) refines report-aligned anatomical details, guided by the rich diagnostic insights recognized from 2D slices.
Then, to map the extracted visual representations to LLM's semantic space, we design Spatial Packer, an efficient projector that compresses 3D visual contexts into a compact set of informative visual tokens.
This projector incorporates a novel Voxel2Point Cross-Attention (V2P-CA), which aggregates high-resolution 3D voxel representations to their centroid points, preserving essential spatial and geometric information.
By integrating spatial packers with the pretrained dual-3D vision encoders, HSENet can effectively capture and transfer hybrid visual representations encompassing both global volumes and detailed anatomies, thereby enabling more accurate text generation. 
We provide comprehensive evaluations across 3D multi-modal retrieval, report generation, and VQA tasks. The results demonstrate that HSENet outperforms existing methods, achieving the state-of-the-art performance in generating discriminative visual representations and high-quality diagnostic responses.

\section{Related Works}
\textbf{Medical Multi-modal Large Language Models.}
MLLMs have shown promise in vision-language applications within the medical field~\cite{Kim2024MDAgents,Xia2024CARES}.
Early explorations such as LLaVA-Med~\cite{Li2023LLaVAMed}, Med-PaLM~\cite{Tu2023Towards}, Flamingo-CXR~\cite{tanno2025collaboration}, and HuatuoGPT-Vision~\cite{Chen2024HuatuoGPTVision} integrate LLMs with 2D medical image encoders for diagnostic reasoning and achieve notable results.
Building on this progress, RadFM~\cite{Wu23RadFM}, M3D~\cite{Bai24M3D}, and CT-CHAT~\cite{Hamamci2024DevelopingGF} extend MLLMs to 3D volumetric data, adapting them for various tasks, e.g., image-text retrieval, report generation, and VQA.
However, these 3D foundational models rely on generic MLLM architectures that struggle to associate intricate 3D structures with medical language, resulting in hallucinations and factual errors. 
To address this, recent studies utilize advanced visual-language alignment strategies, including efficient pretraining~\cite{Xin2025Med3DVLMAE, Blankemeier2024Merlin}, knowledge injection~\cite{Wu2024XLIP, Lai2025Bridged,Ni2024MG3D,Blankemeier2024Merlin}, and dedicated multi-modal projectors~\cite{Shi24Med2E3, Xin2025Med3DVLMAE, Chen2024Dia, Chen2023MedBLIP}.
Unlike the above methods, we introduce a hybrid visual perception and projection pipeline to distill enriched spatial patterns of global volume and local anatomy, enabling accurate and robust 3D vision-language understanding.

\textbf{3D Medical Vision-Language Alignment.}
In medical MLLMs, learning aligned volume and report representations is essential for 3D downstream tasks~\cite{Blankemeier2024Merlin}.
Existing approaches can be broadly categorized into two stages:
\textbf{\textit{1) Vision-language pre-training.}}
\citet{Xin2025Med3DVLMAE} leverages DCFormer~\cite{Gorkem2025DCFormer} and pairwise sigmoid loss~\cite{Xiaohua2023Sigmoid} to achieve efficient yet rich visual-textual alignment.
Besides, additional supervision from external knowledge, such as electronic health records~\cite{Blankemeier2024Merlin}, medical entities~\cite{Wu2024XLIP}, and LLM-summarized text~\cite{Lai2025Bridged} has also been shown to improve alignment quality. 
Nonetheless, abundant patient data is often difficult to obtain, while LLM-generated text may not always be reliable.
In contrast, we leverage informative and readily accessible 2D slices from 3D volumes to promote vision-language consistency and strengthen 3D visual perception.
\textbf{\textit{2) Multi-modal projection in MLLM fine-tuning.}}
\citet{Bai24M3D} compress 3D tokens via spatial pooling to fit LLM input constraints, at the cost of losing spatial details.
Med3DVLM~\cite{Xin2025Med3DVLMAE} integrates MLP-Mixer~\cite{Ilya2021MLPMixer} to capture hierarchical features and improve cross-modal interaction.
Med-2E3~\cite{Shi24Med2E3} projects both 2D slices and 3D volume features directly extracted from frozen encoders to the LLM, but may suffer from inconsistencies between 2D and 3D representations.
In contrast, our approach decouples visual perception and projection processes. By utilizing spatial packers to independently project the visual contexts perceived by our pretrained, correlated dual visual encoders, we produce compact yet expressive hybrid representations that more effectively guide the LLM for clinical reasoning.

\section{Methodology}
\begin{figure*}[t]
\centering  
\includegraphics[width=0.92\linewidth]{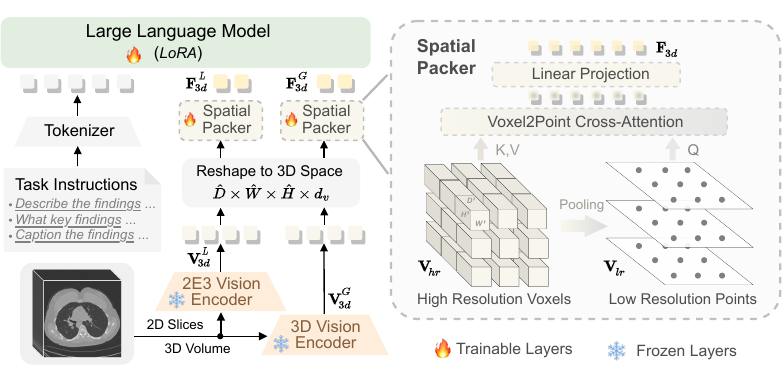}  
\caption{
Architecture of the proposed HSENet. 
The input 3D CT volume is processed in parallel by the 3D Vision Encoder and the 2E3 Vision Encoder to extract rich, multi-scale features.
These hybrid visual representations are then projected by two dedicated spatial packers into the semantic space of LLM, enabling effective 3D medical vision-language modeling.
}
\label{vlm_finetunning}  
\end{figure*}
\subsection{Overview}
Given an input 3D CT volume 
$\mathbf{I}_{3d}\in \mathbb{R} ^{D \times W \times H\times C}$, where $D$, $W$, $H$, and $C$ represent the depth, width, height, and channel of the processed volume, respectively, 
our HSENet aims to learn rich visual representations and prompt the language model to generate the corresponding CT report $R =\{r_1, ..., r_M\}$ with $M$ words.
The architecture of HSENet is shown in Figure~\ref{vlm_finetunning}, which contains the encoding and projecting of hybrid visual features for accurate language generation.

\textbf{Hybrid Visual Encoding.} Clinically, the interpretation of 3D CT scans relies on both macro and micro levels of diagnosis, requiring observations of overall structures and detailed anatomical features~\cite{Liang2009Macro}.
Motivated by this, we introduce dual vision encoders to capture essential 3D medical information:
a 3D Vision Encoder $\mathbf{E}_{\mathrm{3d}}(\cdot)$ for learning global volumetric structures, and a 2E3 Vision Encoder $\mathbf{E}_{\mathrm{2e3}}(\cdot)$ for learning local anatomical features.
These encoders operate in parallel, extracting 3D features $\mathbf{I}_{3d}$, and generating global volumetric features $\mathbf{V}^{G}_{3d} \in \mathbb{R} ^{N_p \times d_v}$ and local anatomical features $\mathbf{V}^{L}_{3d} \in \mathbb{R} ^{N_p \times d_v}$, respectively.
$N_p = (\hat{D} \times \hat{W} \times \hat{H})$ denotes the number of encoded 3D patches, $(\hat{D},\hat{W},\hat{H})$ is the encoded spatial dimensions, and $d_v$ is the feature dimension.

\textbf{Multi-modal Projection.}
To effectively bridge the gap between 3D medical images and the LLM’s semantic space, we introduce a spatial packer that condenses high-resolution 3D regions into a compact set of visual tokens. 
Specifically, we employ twin spatial packers to process global ($\mathbf{V}^{G}_{3d}$) and local ($\mathbf{V}^{L}_{3d}$) 3D visual features in parallel, resulting in transformed features $\mathbf{F}^{G}_{3d} \in \mathbb{R} ^{N_{p}^{'} \times d_t}$ and $\mathbf{F}^{L}_{3d} \in \mathbb{R} ^{N_{p}^{'} \times d_t}$.
Here, $N_{p}^{'}$ denotes the number of compressed tokens, $d_t$ is LLM's feature dimension. 

\textbf{Language Decoding.}
We construct multi-modal prompts by concatenating the projected hybrid visual representations with task instructions, guiding the LLM to generate diagnostic answers.
To optimize the LLM, we employ LoRA~\cite{hu2021lora} and minimize the following cross-entropy loss:
\begin{equation}
\mathcal{L}_{Gen} = - \sum_{t=1}^M \log P(y_t \mid y_{1:t-1}, \{\mathbf{F}^{G}_{3d},\mathbf{F}^{L}_{3d}\}; \theta),
\end{equation}
where $P(y_t|*)$ denotes the probability of predicting text token $y_t$ conditioned on the preceding tokens $y_{1:t-1}$ and the projected hybrid visual features $\mathbf{F}^{G}_{3d}$ and $\mathbf{F}^{L}_{3d}$.
$\theta$ denotes the trainable parameters.

\begin{figure*}[t]
\centering  
\includegraphics[width=0.92\linewidth]{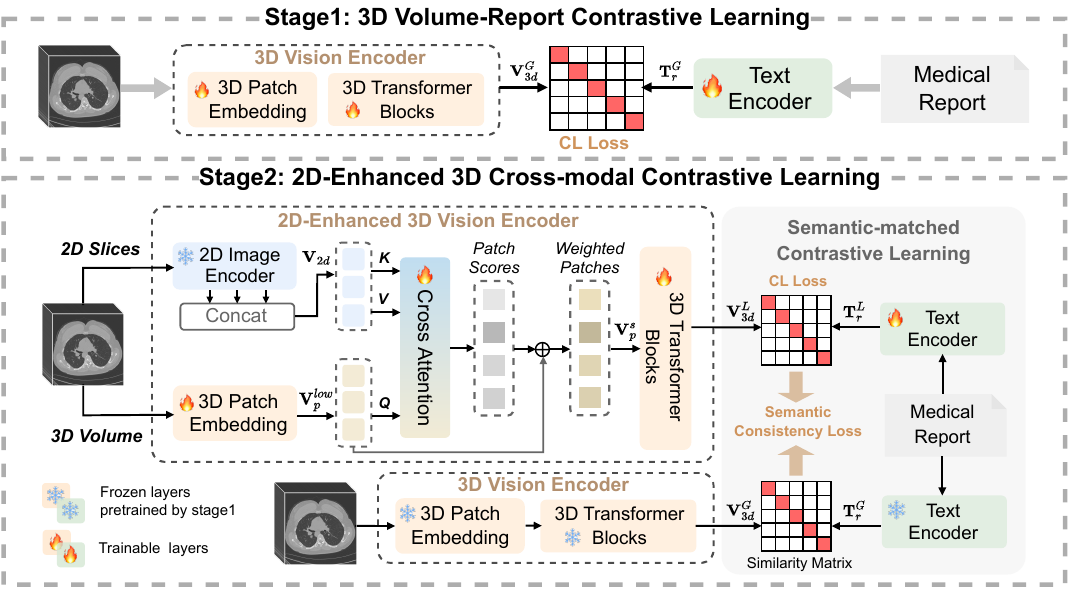}  
\caption{
Overview of the dual-stage pretraining framework.
\textbf{Stage 1}: The 3D Vision Encoder is trained for global vision-language alignment using paired 3D volumes and medical reports.
\textbf{Stage 2}: The 2E3 Vision Encoder is trained to exploit anatomy-related local 3D patches aligned with reports. 
A semantic consistency loss is applied in Stage 2 to maintain alignment with the global relations learned in Stage 1, ensuring a stable local representation refining.
}
\label{clip_pretraining}  
\end{figure*}

\subsection{Dual-stage 3D Medical Vision-Language Pretraining}
\label{method:dual_pretraining}
To mimic the way physicians observe macro and micro 3D visual patterns, we design a novel dual-stage cross-modal pretraining framework to build robust 3D vision encoders.
As illustrated in Figure~\ref{clip_pretraining}, we first conduct 3D volume-report contrastive learning to train a 3D Vision Encoder for capturing macro-level CT structures.
Then, we propose 2D-enhanced 3D (2E3) cross-modal contrastive learning to refine the 2E3 Vision Encoder by incorporating detailed 3D anatomical patterns, leveraging cross-modal relations, enriched semantics from related 2D slices.

\textbf{Stage 1: 3D Volume-Report Contrastive Learning.}
We harness expert-written reports as inherent labels to learn discriminative visual representations of 3D CT volumes.
Following common paradigms~\cite{Radford2021CLIP,Zhang2024BiomedCLIPAM}, we pair a 3D vision encoder $\mathbf{E}_{\mathrm{3d}}(\cdot)$ and a text encoder $\mathbf{E}_{text}^{s_1}(\cdot)$ to extract volume features $\mathbf{V}^{G}_{3d}$ and report features $\mathbf{T}^{G}_r$, respectively.
To align these features, we leverage the CLS token from each encoder as a compact summary embedding, which is then projected into a shared latent space $\tilde{\mathbf{x}}_{3d} \in \mathbb{R} ^{d_l}$ and $\tilde{\mathbf{x}}_{t} \in \mathbb{R} ^{d_l}$.
The objective of this stage is to maximize the mutual information between paired volume and report, achieved by optimizing symmetric InfoNCE~\cite{oord2018representation} loss:
\begin{equation}
\label{method:stage1_cl_loss}
\mathcal{L}_{CL} = -\frac{1}{2N_c} \sum_{i=1}^{N_c} \left( \log \frac{\exp(\text{sim}(\tilde{\mathbf{x}}_{3d}^{(i)}, \tilde{\mathbf{x}}_{t}^{(i)})/\tau)}{\sum_{k=1}^{B} \exp(\text{sim}(\tilde{\mathbf{x}}_{3d}^{(i)}, \tilde{\mathbf{x}}_{t}^{(k)})/\tau)} + \log \frac{\exp(\text{sim}(\tilde{\mathbf{x}}_{t}^{(i)}, \tilde{\mathbf{x}}_{3d}^{(i)})/\tau)}{\sum_{k=1}^{B} \exp(\text{sim}(\tilde{\mathbf{x}}_{t}^{(i)}, \tilde{\mathbf{x}}_{3d}^{(k)})/\tau)} \right),
\end{equation}
where $sim(\cdot)$ computes the cosine similarity, $B$ is the batch size, $\tau$ is the temperature hyperparameter, and $N_{c}$ denotes the number of volume-report pairs.

\textbf{Stage 2: 2D-Enhanced 3D Cross-modal Contrastive Learning.}
Radiologists are skilled in correlating 3D contextual information with 2D slice-level observations to interpret subtle anatomies~\cite{Salvolini2000Clinical}. 
Inspired by this, we distill knowledge in 2D slices to refine 3D vision-language alignment from global to fine-grained anatomy. 
This approach is more promising than current methods that rely on external patient data~\cite{Wu2024XLIP}, which is often inaccessible,
or on potentially unreliable LLM-generated texts~\cite{Lai2025Bridged}, since our 2D slices can be readily obtained from 3D volumes and inherently contain rich diagnostic information.

\textbf{\textit{2D Slice Processing}.} 
We uniformly slice the 3D volume along the Z-axis and obtain $\mathbf{I}_{2d}=\{s_1, s_2, ..., s_{N_s}\}$, where $N_s$ represents the number of extracted slices.
We extract slice features by processing each slice with pre-trained BioMedCLIP~\cite{Zhang2024BiomedCLIPAM}, then stacking them into $\mathbf{V}_{2d}\in \mathbb{R} ^{N_s \times d_v}$.

\textbf{\textit{2D-Enhanced 3D Vision Enhancing}.}
Unlike previous methods that focus on augmenting high-level 3D visual features~\cite{Lai2025Bridged,Wu2024XLIP}, we argue that low-level features carry richer 3D spatial cues for capturing anatomical details and improving visual representation quality.
Accordingly, as illustrated in the bottom of Figure~\ref{clip_pretraining}, we introduce a 2D-enhanced 3D vision encoder $\mathbf{E}_{\mathrm{2e3}}(\cdot)$ for local vision enhancement.
Firstly, we extract low-level 3D patch features $\mathbf{V}^{low}_{p}\in \mathbb{R} ^{N_p \times d_v}$ from the 3D patch embedding layer of a standard 3D ViT.
We then interact $\mathbf{V}^{low}_{p}$ with 2D features $\mathbf{V}_{2d}$ by cross-attention layers, to estimate the significance of distinct 3D patches:
\begin{equation}
\mathbf{S}_{3d} = FFN(MHA(\mathbf{V}^{low}_{p}, \mathbf{V}_{2d}, \mathbf{V}_{2d}),
\end{equation}  
where \textit{FFN} and \textit{MHA} denote feed-forward and multi-head attention layers. 
The resulting scoring feature $\mathbf{S}_{3d}\in \mathbb{R}^{N_p\times d}$ is then projected via MLP layers to produce patch scores $\mathbf{S}^{'}_{3d}=\{s^{(1)}_{3d}, s^{(2)}_{3d}, ..., s^{(N_p)}_{3d}\}\in \mathbb{R}^{N_p}$, with $s^{(i)}_{3d}$ indicating the importance of the i-th 3D patch.
Using these scores, we weight the low-level 3D patch features, yielding $\mathbf{V}_{p}^{s}\in \mathbb{R} ^{N_p \times d_v}$, which emphasizes diagnostically relevant spatial areas.
Finally, $\mathbf{V}_{p}^{s}$ is fed through transformer blocks to generate high-level vision features $\mathbf{V}_{3d}^{L}$ that capture local 3D anatomical details.

\textbf{\textit{Semantic-Matched Contrastive Learning}.}
To capture fine-grained anatomical representations, we apply contrastive learning loss $\mathcal{L}_{CL}^{2e3}$ similar to Equation~\ref{method:stage1_cl_loss}, aligning the enhanced 3D visual features $\mathbf{V}_{3d}^{L}$ with the corresponding report features produced by text decoder $\mathbf{E}_{text}^{s_2}(\cdot)$.  
While this objective encourages detailed local alignment, unconstrained optimization risks drifting from generalizable vision-text relationships.
To mitigate this, we introduce a semantic consistency loss $\mathcal{L}_{SA}$ that regularizes the cross-modal similarity matrix by anchoring it to the global alignment established in Stage 1. The loss function is formulated as:
\begin{equation}
    \mathcal{L}_{SA} = \sum_{i=1}^{B} \left\|\text{sim}(\tilde{\mathbf{x}}_{3d}^{(i)}, \tilde{\mathbf{x}}_{t1}^{(i)})/\tau)-\text{sim}(\tilde{\mathbf{x}}_{2e3}^{(i)}, \tilde{\mathbf{x}}_{t2}^{(i)})/\tau)\right\|^2,
\end{equation}
where $\tilde{\mathbf{x}}_{3d}$, $\tilde{\mathbf{x}}_{t1}$ are the volume and report features from the fixed Stage 1 encoders $\mathbf{E}_{\mathrm{3d}}(\cdot)$ and $\mathbf{E}_{text}^{s_1}(\cdot)$, respectively. 
$\tilde{\mathbf{x}}_{2e3}$ and $\tilde{\mathbf{x}}_{t2}$ are the local vision features and report features from stage~2.
The overall loss in stage 2 can be calculated as:
\begin{equation}
\mathcal{L}_{SCL} = \mathcal{L}_{CL}^{2e3} + \lambda_s \mathcal{L}_{SA},\label{stage_final_loss},
\end{equation}
where $\lambda_s$ controls the regularization strength.
During Stage 2, the Stage 1 encoders are frozen, while $\mathbf{E}_{\mathrm{2e3}}(\cdot)$ and $\mathbf{E}_{text}^{s_2}(\cdot)$ are trainable.
This formulation preserves foundational knowledge from Stage~1 while refining representations in Stage 2, enhancing the model’s capacity to capture fine-grained anatomical details and maintain robust vision-text alignment.

\subsection{Spatial Packer}
As shown in Figure~\ref{vlm_finetunning}, we propose spatial packers to project the extracted global and local 3D visual features ($\mathbf{V}_{3d}^{G}$ and $\mathbf{V}_{3d}^{L}$) into LLM's latent space.
The key insight behind spatial packer is to leverage both high- and low-resolution embeddings for efficient token compression and spatial preservation.
Here, we illustrate the workflow of spatial packer using $\mathbf{V}_{3d}^{G}$ as a representative example.

\textbf{High-Resolution Voxel Embedding.}
Following~\citet{Bai24M3D}, we reshape the patch dimension $N_p$ of  $\mathbf{V}_{3d}^{G}\in \mathbb{R}^{N_p \times d_v}$ back to its original 3D spatial layout, obtaining $\mathbf{V}_{3d'}^{G}\in \mathbb{R}^{\hat{D} \times \hat{W} \times \hat{H} \times d_v}$.
We then partition $\mathbf{V}_{3d'}^{G}$ along each spatial axis using strides ($S_d$,$S_w$,$S_h$), resulting in high-resolution voxel features
$\mathbf{V}_{hr}^{G}\in \mathbb{R}^{(S_d\cdot S_w\cdot S_h) \times D' \times W' \times  H' \times d_v}$,
where $D' = \frac{\hat{D}}{S_d}$, $W' = \frac{\hat{W}}{S_w}$, and $H' = \frac{\hat{H}}{S_h}$ denotes the spatial dimensions of each local voxel (see right part of Figure~\ref{vlm_finetunning}).
$\mathbf{V}_{hr_{i,j,k}}^{G}\in \mathbb{R}^{D' \times W' \times H' \times d_v}$ represents the spatial feature of the voxel coordinated at ($i$, $j$, $k$) in volume space.

\textbf{Low-Resolution Point Embedding.}
To capture the overall pattern within each local voxel, we apply feature pooling for $\mathbf{V}_{hr_{i,j,k}}^{G}\in \mathbb{R}^{D' \times W' \times H' \times d_v}$, extracting a centroid point representation $\mathbf{V}_{lr_{i,j,k}}^{G} \in \mathbb{R}^{d_v}$.
For the entire 3D volume, these centroid embeddings aggregated into the low-resolution point embedding $\mathbf{V}_{lr}^{G}\in \mathbb{R}^{S_d \times S_w \times S_h \times d_v}$, where $S_d$, $S_w$, and $S_h$ denote the number of points along each spatial dimension.

\textbf{Voxel2Point Cross-Attention.}
We propose a Voxel2Point Cross-Attention (V2P-CA) mechanism to inject enriched spatial clues from high-resolution $\mathbf{V}_{hr}^{G}$ into low-dimensional $\mathbf{V}_{lr}^{G}$, enabling efficient visual projection.
Unlike previous cross-attention-based projectors~\cite{Li2024Mini,Li2024TokenPacker,Chen2024Dragonfly} that are limited to 2D images, our V2P-CA learns 3D voxel-point interactions for effective spatial preservation.
We first reshape $\mathbf{V}_{lr}^{G}$ as low-resolution query  $Q_{l}\in \mathbb{R}^{(S_d \cdot S_w \cdot S_h) \times 1 \times d_v}$, and reshape $\mathbf{V}_{hr}^{G}$ as high-resolution key $K_{h}\in \mathbb{R}^{(S_d\cdot S_w\cdot S_h) \times (D' \cdot W' \cdot  H') \times d_v}$ and value $V_{h}\in \mathbb{R}^{(S_d\cdot S_w\cdot S_h) \times (D' \cdot W' \cdot  H') \times d_v}$.
Then, we leverage cross-attention to make each point in $Q_{l}$ fully absorb its corresponding fine-grained voxel features in $K_{h}$ and $V_{h}$:
\begin{equation}
    \mathbf{Y}_{3d}^{G} = FFN(MHA(Q_{l}, K_{h}, V_{h})),
\end{equation}
where $\mathbf{Y}_{3d}^{G}\in \mathbb{R}^{(S_d \cdot S_w \cdot S_h) \times d_v}$ denotes the compact spatial visual tokens. 
We finally use 2-layer MLPs to map the $\mathbf{Y}_{3d}^{G}$ to LLM's latent dimension, producing $\mathbf{F}_{3d}^{G}\in \mathbb{R}^{N_{p}^{'} \times d_t}(N_{P}^{'} = S_d \cdot S_w \cdot S_h)$.
We adopt the same procedure to generate $\mathbf{F}_{3d}^{L}$ for local anatomical features $\mathbf{V}_{3d}^{L}$.

\section{Experiments and Results}
\subsection{Experiment Settings}
\textbf{Tasks and Datasets.}
To validate HSENet for 3D medical vision-language understanding, we evaluate on three tasks: (1) medical image-text retrieval, (2) report generation, and (3) medical VQA.
For image-text retrieval and report generation tasks, we use the benchmark 3D CT dataset CT-RATE~\cite{Hamamci2024DevelopingGF}, which contains 25,692 non-contrast chest CT scans from 21,304 anonymized patients. 
After data expansion and excluding cases with excessively short or invalid reports, we retain 47,149 volume-report pairs (20,000 unique patients) for training and 3,039 pairs (1,304 distinct patients) for testing.
For medical VQA, we adopt the RadGenome-ChestCT dataset~\cite{Zhang2024RadGenome}, which contains 302,827 open-ended VQA pairs focused on 3D location observations, enabling the evaluation of models' 3D spatial reasoning capabilities.
We allocate 285,086 samples for training and 17,741 samples for testing.

\textbf{Implementation Details.}
We employ the standard 3D Vision Transformer (3D ViT)~\cite{Dosovitskiy20213dViT} and Bert~\cite{Devlin2019BERT} as visual and language encoders for pretraining and retrieval.
We utilize Phi4-4B-Instruct~\cite{Abouelenin2025Phi4} as the language model, which is integrated with our pretrained dual visual encoders and spatial packers to construct MLLM.
Following~\citet{Bai24M3D}, input volumes are normalized and resized to $(D,W,H)=(32,256,256)$
using Min-Max Normalization, then encoded into patches of size $(\hat{D},\hat{W},\hat{H})=(8,16,16)$ by 3D ViT.
The spatial packer uses strides $(S_d,S_w,S_h)=(8,4,4)$, yielding local voxels of size $(D',W',H')=(1,4,4)$.
We set the number of 2D slices $N_s=32$, loss weight $\lambda_s=0.1$, and feature dimensions $d_v=768$, $d_l=512$, $d_t=3072$.
Experiments are conducted on 8 RTX 3090 GPUs using AdamW optimizer. Both pretraining stages run for 50 epochs with a learning rate of 1e-4.
Report generation is trained for 6 epochs at 1e-4 and VQA for 4 epochs at 5e-5.
Additional details are provided in the supplementary material.

\textbf{Evaluation Metrics.}
We use Recall@K (R@5/10/50/100) to evaluate top-k retrieval accuracy in report-to-volume and volume-to-report tasks.
For volume-to-volume retrieval, we utilize Mean Average Precision (MAP@5/10/50) to assess the model’s ability to retrieve pathology-relevant volumes.
Report generation is evaluated using standard natural language generation (NLG) metrics (BLEU~\cite{papineni2002bleu}, ROUGE~\cite{lin2004rouge}, METEOR~\cite{lavie2007meteor}, and BERTScore~\cite{Zhang2020BERTScore}) to measure linguistic quality, along with RaTE-Score~\cite{Zhao2024RaTEScore} to assess clinical relevance.
For the VQA task, we use both the NLG metrics and answer accuracy.
The accuracy evaluates the performance separately on major (e.g., lung, heart) and minor (e.g., left lung lower lobe, left heart ventricle) location categories.

\begin{table}[t]
\centering
\scriptsize  
\caption{
Experiments on image-text retrieval performance.
\textbf{Bold} indicates the best performance, while \underline{underlined} indicates the second-best performance for each model. 
\textit{3D-ViT} and \textit{2E3-ViT} refer to our 3D Vision Encoder $\mathbf{E}_{\mathrm{3d}}(\cdot)$ and 2E3 Vision Encoder $\mathbf{E}_{\mathrm{2e3}}(\cdot)$, respectively.
† denotes the model reproduced using the official code.
\textit{Full Text}, \textit{Text CLS}, and \textit{2D Slices} refer to the features used for guiding patch scoring within $\mathbf{E}_{\mathrm{2e3}}(\cdot)$.
}
\label{table:retrieval}
\setlength{\tabcolsep}{4.6pt}{
\begin{tabular}{lccccccccccc}
\toprule
\multicolumn{1}{c}{\multirow{2}{*}{\textbf{Methods}}} &
  \multicolumn{4}{c}{\textbf{Report-to-Volume Retrieval}} &
  \multicolumn{4}{c}{\textbf{Volume-to-Report Retrieval}} &
  \multicolumn{3}{c}{\textbf{Volume-to-Volume Retrieval}} \\ \cline{2-12} \\[-1.5ex]
\multicolumn{1}{c}{} &
  \textbf{R@5} &
  \textbf{R@10} &
  \textbf{R@50} &
  \textbf{R@100} &
  \textbf{R@5} &
  \textbf{R@10} &
  \textbf{R@50} &
  \textbf{R@100} &
  \textbf{MAP@5} &
  \textbf{MAP@10} &
  \textbf{MAP@50} \\ \hline
\multicolumn{12}{l}{\textit{(a) comparison with state-of-the-art pretraining models}} \\
VocabFine\cite{Hamamci2024DevelopingGF} &
  0.10 &
  0.60 &
  2.30 &
  2.00 &
  / &
  / &
  / &
  / &
  68.30 &
  57.20 &
  48.80 \\
MG-3D\cite{Ni2024MG3D} &
  / &
  / &
  3.88 &
  / &
  / &
  / &
  / &
  / &
  / &
  / &
  / \\
Merlin\cite{Blankemeier2024Merlin} &
  1.50 &
  2.70 &
  7.70 &
  12.70 &
  / &
  / &
  / &
  / &
  62.60 &
  51.30 &
  43.90 \\
CT-CLIP\cite{Hamamci2024DevelopingGF} &
  2.90 &
  5.00 &
  18.00 &
  28.70 &
  / &
  / &
  / &
  / &
  68.30 &
  57.20 &
  48.90 \\
M3D-CLIP\cite{Bai24M3D}† &
  4.87 &
  8.72 &
  24.42 &
  33.89 &
  5.30 &
  8.88 &
  24.38 &
  34.16 &
  {\ul 68.80} &
  57.83 &
  49.54 \\
Med3DVLM\cite{Xin2025Med3DVLMAE}† &
  2.96 &
  4.94 &
  15.56 &
  23.89 &
  2.44 &
  3.78 &
  12.41 &
  18.33 &
  68.31 &
  56.98 &
  48.31 \\
\rowcolor[HTML]{EFEFEF} 
Ours (3D-ViT)\rule{0pt}{2.0ex} &
  {\ul 5.76} &
  {\ul 9.28} &
  {\ul 25.50} &
  {\ul 34.72} &
  {\ul 5.63} &
  {\ul 9.05} &
  {\ul 25.67} &
  {\ul 34.62} &
  68.75 &
  {\ul 57.85} &
  {\ul 49.57} \\
\rowcolor[HTML]{EFEFEF} 
Ours (2E3-ViT) &
  \textbf{5.82} &
  \textbf{9.44} &
  \textbf{28.46} &
  \textbf{39.85} &
  \textbf{6.09} &
  \textbf{9.67} &
  \textbf{28.63} &
  \textbf{39.22} &
  \textbf{69.32} &
  \textbf{58.68} &
  \textbf{50.58} \\ \hline\hline
\multicolumn{12}{l}{\textit{(b) diffrernt settings for 3D patch scoring}} \\
Full Text &
  1.61 &
  3.26 &
  10.89 &
  16.72 &
  1.51 &
  2.96 &
  10.53 &
  16.42 &
  66.56 &
  55.17 &
  46.93 \\
Text CLS &
  {\ul 2.93} &
  {\ul 5.76} &
  {\ul 19.32} &
  {\ul 29.48} &
  3.32 &
  {\ul 6.12} &
  {\ul 19.78} &
  {\ul 28.59} &
  {\ul 68.00} &
  {\ul 57.10} &
  {\ul 48.85} \\
\rowcolor[HTML]{EFEFEF} 
2D Slice\rule{0pt}{2.0ex} &
  \textbf{5.82} &
  \textbf{9.44} &
  \textbf{28.46} &
  \textbf{39.85} &
  \textbf{6.09} &
  \textbf{9.67} &
  \textbf{28.63} &
  \textbf{39.22} &
  \textbf{69.32} &
  \textbf{58.68} &
  \textbf{50.58} \\ \hline\hline
\multicolumn{12}{l}{\textit{(c) ablation study of semantic consistency loss}} \\
w/o $\mathcal{L}_{SA}$ &
  {\ul 4.90} &
  {\ul 8.29} &
  {\ul 27.28} &
  {\ul 37.97} &
  {\ul 4.77} &
  {\ul 9.15} &
  {\ul 26.82} &
  {\ul 37.94} &
  {\ul 69.12} &
  {\ul 58.45} &
  {\ul 50.38} \\
\rowcolor[HTML]{EFEFEF} 
Ours (2E3-ViT)\rule{0pt}{2.0ex} &
  \textbf{5.82} &
  \textbf{9.44} &
  \textbf{28.46} &
  \textbf{39.85} &
  \textbf{6.09} &
  \textbf{9.67} &
  \textbf{28.63} &
  \textbf{39.22} &
  \textbf{69.32} &
  \textbf{58.68} &
  \textbf{50.58} \\[-0.6ex]
\bottomrule
\end{tabular}
}
\end{table}

\subsection{Results on Medical Image-Text Retrieval}
To assess the capability of the pretrained dual 3D vision encoders, we evaluate their effectiveness across various retrieval tasks: 1) report-to-volume retrieval, 2) volume-to-report retrieval, and 3) volume-to-volume retrieval.

\begin{wrapfigure}{r}{0.6\linewidth}  
  \centering
  \vspace{-10pt}  
  \includegraphics[width=0.98\linewidth]{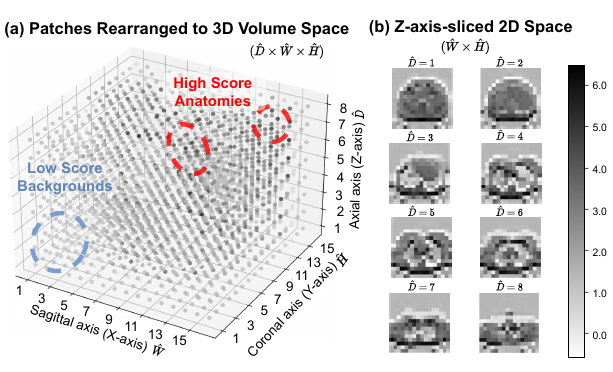}
  \caption{
  Visualization of 3D patch scores in 2E3 Vision Encoder. Darker colors indicate higher scores. 
  (a) Patches are rearranged into the original 3D volume space to illustrate their score distribution. The model assigns higher scores to semantically essential patches (highlighted in \textcolor{red}{\textbf{red}}) and lower scores to less relevant patches (in \textcolor{blue}{\textbf{blue}}).
  (b) Axial slices along the Z-axis reveal the patch scores at different depth levels $\hat{D}$, providing a clearer view of the score variations.
  }
  \label{patch_score}
\end{wrapfigure}
\textbf{Comparisons with 3D Pretraining Models.}
We compare the performance with state-of-the-art 3D medical vision-language pretraining models,
including CT-CLIP~\cite{Hamamci2024DevelopingGF}, M3D-CLIP~\cite{Bai24M3D}, VocabFine~\cite{Hamamci2024DevelopingGF}, MG-3D~\cite{Ni2024MG3D}, Merlin~\cite{Blankemeier2024Merlin}, and Med3DVLM~\cite{Xin2025Med3DVLMAE}.
As evidenced by Table~\ref{table:retrieval}(a), our method achieves consistent superiority across all evaluation metrics in three retrieval tasks. 
Notably, our 3D Vision Encoder, based on simple 3D patch processing, achieves 1.99× higher R@5 in report-to-volume retrieval compared to the more complex hierarchical volume partitioning and encoding pipeline in CT-CLIP 
(5.76\% vs. 2.90\%).
By incorporating 2D slice-guided patch scoring, our 2E3 Visual Encoder yields substantial gains, improving R@100 by $\sim5\%$ over the 3D Vision Encoder. 
This indicates that learning local anatomical patterns from 2D slices can effectively model the intricate 3D volume-report relations.
The 2E3 Vision Encoder also achieves SoTA volume-to-volume retrieval performance, suggesting that our model learns discriminative 3D medical features, thereby retrieving volumes with high pathological relevance.

\textbf{Ablation Studies.}
To evaluate the effectiveness of our 2D-guided patch scoring, we replaced 2D slice features with alternative text-derived features, including the full text embedding $\mathbf{T}_{full} \in \mathbb{R}^{512 \times 768}$ and the CLS token $\mathbf{T}_{cls} \in \mathbb{R}^{768}$ from the text encoder $\mathbf{E}_{text}^{s_2}(\cdot)$.
As shown in Table~\ref{table:retrieval}(b), both variants lead to performance drops, particularly with $\mathbf{T}_{full}$ (R@5: from 5.82\% to 1.61\%, 4.21\%$\downarrow$).
This may be due to the inherent gap between high-level textual features and our low-level visual patches, which prevents visual enhancement as achieved by 2D slices. 
Additionally, removing the semantic consistency loss $\mathcal{L}_{SA}$ also results in performance degradation (see Table~\ref{table:retrieval}(c)).
This confirms its importance in maintaining stable cross-modal correspondence during local feature refinement.

\begin{table}[t]
\centering
\scriptsize  
\caption{
Experiments on report generation.
\textbf{Bold} indicates the best performance, while \underline{underlined} indicates the second-best performance. 
† denotes the model reproduced using the official code.
}
\label{table:report_generation}
\setlength{\tabcolsep}{3.5pt}{
\begin{tabular}{lccccccccc}
\toprule
\multicolumn{1}{c}{\textbf{Methods}} &
  \textbf{BLEU-1} &
  \textbf{BLEU-2} &
  \textbf{BLEU-3} &
  \textbf{BLEU-4} &
  \textbf{ROUGE-1} &
  \textbf{ROUGE-L} &
  \textbf{METEOR} &
  \textbf{BERT-Score} &
  \textbf{RaTE-Score} \\ \hline
\multicolumn{10}{l}{\textit{(a) comparison with state-of-the-art models}}                            \\
RadFM\cite{Wu23RadFM}            & 29.85          & /              & /              & /              & 45.67          & /              & 28.75          & 86.97          & /              \\
CT-CHAT\cite{Hamamci2024DevelopingGF}          & 39.52          & /              & /              & /              & /              & 32.12          & 21.85          & /              & /              \\
M3D-LaMed\cite{Bai24M3D}        & 40.32          & /              & /              & /              & 52.08          & /              & 36.67          & 87.55          & /              \\
E3D-GPT\cite{Lai2024E3D}          & 41.15          & /              & /              & /              & 52.60          & /              & 41.79          & 87.97          & /              \\
Med-2E3\cite{Shi24Med2E3}†         & 55.87          & 30.82          & 19.64          & 14.09          & {\ul 54.40}    & 33.33          & 43.06          & 87.99          & {\ul 61.81}    \\
Med3DVLM\cite{Xin2025Med3DVLMAE}†        & {\ul 56.76}    & {\ul 32.20}    & 21.46          & {\ul 16.00}    & 54.38          & {\ul 34.17}    & {\ul 43.18}    & {\ul 88.12}    & 61.07          \\
\rowcolor[HTML]{EFEFEF} 
HSENet (Ours)\rule{0pt}{2.0ex}    & \textbf{62.89} & \textbf{39.47} & \textbf{29.11} & \textbf{24.01} & \textbf{56.50} & \textbf{40.63} & \textbf{44.75} & \textbf{88.99} & \textbf{64.99} \\ \hline\hline
\multicolumn{10}{l}{\textit{(b) comparison of different multi-modal projectors utilized in HSENet}}                                                                                                   \\
Q-Former\cite{Li2023BLIP2}         & 55.60          & 32.30          & 22.15          & 17.11          & 53.62          & 35.47          & 43.29          & 87.97          & 62.39          \\
Sequence Pooling\cite{Bai24M3D} & 56.20          & 33.40          & 23.40          & 18.46          & 53.61          & 36.29          & 43.51          & 88.08          & 62.82          \\
Spatial Pooling\cite{Bai24M3D}  & {\ul 61.67}    & {\ul 37.20}    & {\ul 26.14}    & {\ul 20.66}    & {\ul 56.48}    & {\ul 38.54}    & {\ul 44.21}    & {\ul 88.84}    & {\ul 63.16}    \\
\rowcolor[HTML]{EFEFEF} 
Spatial Packer\rule{0pt}{2.0ex}   & \textbf{62.89} & \textbf{39.47} & \textbf{29.11} & \textbf{24.01} & \textbf{56.50} & \textbf{40.63} & \textbf{44.75} & \textbf{88.99} & \textbf{64.99} \\[-0.6ex]
\bottomrule
\end{tabular}
}
\end{table}

\begin{table}[t]
\centering
\scriptsize  
\caption{
Ablation studies on different settings of the visual encoders and the projector in medical report generation. \textit{3D-ViT} and \textit{2E3-ViT} denotes the proposed 3D Visual Encoder $\mathbf{E}_{\mathrm{3d}}(\cdot)$ and 2E3 Visual Encoder $\mathbf{E}_{\mathrm{2e3}}(\cdot)$, respectively.
}
\label{table:ablation_report_generation}
\setlength{\tabcolsep}{1.27pt}{
\begin{tabular}{ccccccccccccc}
\toprule
\multirow{2}{*}{\textbf{Methods}} &
  \multicolumn{2}{c}{\textbf{Vision Encoder}} &
  \multirow{2}{*}{\textbf{\begin{tabular}[c]{@{}c@{}}Spatial\\ Packer\end{tabular}}} &
  \multirow{2}{*}{\textbf{BLEU-1}} &
  \multirow{2}{*}{\textbf{BLEU-2}} &
  \multirow{2}{*}{\textbf{BLEU-3}} &
  \multirow{2}{*}{\textbf{BLEU-4}} &
  \multirow{2}{*}{\textbf{ROUGE-1}} &
  \multirow{2}{*}{\textbf{ROUGE-L}} &
  \multirow{2}{*}{\textbf{METEOR}} &
  \multirow{2}{*}{\textbf{BERT-Score}} &
  \multirow{2}{*}{\textbf{RaTE-Score}} \\ \cline{2-3} \\[-2.0ex]
 &
  \textbf{3D-ViT} &
  \textbf{2E3-ViT} &
   &
   &
   &
   &
   &
   &
   &
   &
   &
   \\ \hline \\[-2.0ex]
(a) &
  $\checkmark$ &
  {\ul } &
   &
  55.57 &
  32.81 &
  22.88 &
  17.93 &
  53.16 &
  35.86 &
  43.14 &
  87.96 &
  62.35 \\
(b) &
  $\checkmark$ &
   &
  $\checkmark$ &
  58.69 &
  34.24 &
  23.43 &
  17.87 &
  55.06 &
  35.84 &
  43.59 &
  88.34 &
  62.74 \\
(c) &
   &
  $\checkmark$ &
   &
  57.87 &
  33.81 &
  23.17 &
  17.79 &
  54.86 &
  35.92 &
  43.69 &
  88.25 &
  62.84 \\
(d) &
   &
  $\checkmark$ &
  $\checkmark$ &
  60.96 &
  36.37 &
  25.44 &
  19.95 &
  56.06 &
  37.65 &
  43.99 &
  88.69 &
  {\ul 63.37} \\
(e) &
  $\checkmark$ &
  $\checkmark$ &
   &
  {\ul 61.67} &
  {\ul 37.20} &
  {\ul 26.14} &
  {\ul 20.66} &
  {\ul 56.48} &
  {\ul 38.54} &
  {\ul 44.21} &
  {\ul 88.84} &
  63.16 \\
\rowcolor[HTML]{EFEFEF} 
Ours\rule{0pt}{2.0ex} &
  $\checkmark$ &
  $\checkmark$ &
  $\checkmark$ &
  \textbf{62.89} &
  \textbf{39.47} &
  \textbf{29.11} &
  \textbf{24.01} &
  \textbf{56.50} &
  \textbf{40.63} &
  \textbf{44.75} &
  \textbf{88.99} &
  \textbf{64.99} \\ [-0.6ex]
  \bottomrule
\end{tabular}
}
\end{table}

\begin{wraptable}{r}{0.75\textwidth}  
\centering
\vspace{-10pt}  
\scriptsize  
\caption{
Experiments on 3D medical VQA. \textit{Major Class Acc} measures the accuracy in answering the major location category, while \textit{Minor Class Acc} evaluates accuracy on more detailed body locations. 
}
\vspace{10pt}
\label{table:vqa}
\setlength{\tabcolsep}{1.0pt}{
\begin{tabular}{lcccccc}
\toprule
\multicolumn{1}{c}{\textbf{Methods}} & \textbf{BLEU-1} & \textbf{ROUGE-1} & \textbf{METEOR} & \textbf{BERT-Score} & \textbf{Major Class Acc.} & \textbf{Minor Class Acc.} \\ \hline
M3D-LaMed\cite{Bai24M3D}†           & 60.15       & 56.49 & 21.25 & 90.36 & 71.61 & 28.08 \\
Med-2E3\cite{Shi24Med2E3}†             & 63.45       & 52.36 & 17.44 & 89.84 & 66.70 & 25.71 \\
Med3DVLM\cite{Xin2025Med3DVLMAE}†            & {\ul 64.11} & 57.08 & 19.59 & 90.65 & 70.39 & 28.90 \\ \hline
Ours (3D-ViT) & 59.70       & 56.80 & 21.66 & 90.56 & 71.07 & 28.84 \\
Ours (2E3-ViT)                & 61.17           & {\ul 57.85}      & {\ul 21.74}     & {\ul 90.71}         & {\ul 72.28}               & {\ul 29.59}               \\
\rowcolor[HTML]{EFEFEF} 
Ours (Dual-ViTs)\rule{0pt}{2.0ex}             & \textbf{65.65}  & \textbf{58.90}   & \textbf{21.58}  & \textbf{90.77}      & \textbf{73.60}            & \textbf{30.28}            \\ [-0.6ex]\bottomrule
\end{tabular}
}
\end{wraptable}
\textbf{Visualization of 3D Patch Scores.} 
Figure~\ref{patch_score} visualizes the 3D patch scores produced by the 2E3 Vision Encoder, demonstrating its capacity to differentiate informative regions from irrelevant ones.
Both volumetric (3D) and slice-based (2D) views are presented to show the spatial distribution of scores.
The model consistently assigns higher scores to anatomically salient patches, thereby enhancing the effectiveness of local representation learning.

\subsection{Results on Medical Report Generation}
\textbf{Comparison Studies.} 
We compare the report generation performance with advanced 3D medical MLLMs, including RadFM~\cite{Wu23RadFM}, CT-CHAT~\cite{Hamamci2024DevelopingGF}, M3D-LaMed~\cite{Bai24M3D}, E3D-GPT~\cite{Lai2024E3D}, Med-2E3~\cite{Shi24Med2E3}, and Med3DVLM~\cite{Xin2025Med3DVLMAE}.
As shown in Table~\ref{table:report_generation}(a), our model achieves state-of-the-art performance across all NLG metrics and the RaTE-Score, reflecting both linguistic fluency and clinical accuracy.
Foundation models such as RadFM, CT-CHAT, and M3D-LaMed mainly adopt generic MLLM architectures and lack dedicated designs to grasp 3D medical clues, leading to lower overall scores.
Med3DVLM uses DCFormer for multi-scale volumetric features and improves BLEU scores, while Med-2E3 enhances clinical relevance (RaTE-Score +0.74\%) by fusing 2D and 3D features for LLM inference, though sacrificing coherence (only 14.09\% of BLEU-4).
In contrast, our method effectively decouples the 3D perception and projection, yielding superior overall results.

\textbf{Ablation Studies.}
We perform ablation studies to assess the impact of the spatial packer and dual visual encoders. 
As shown in Table~\ref{table:report_generation}, replacing our spatial packer with Q-Former or pooling strategies degrades performance, with Q-Former leading to a 7.29\% BLEU-1 drop, likely due to disrupted 3D structure.
Table~\ref{table:ablation_report_generation} compares different visual encoder configurations: Results from settings (a), (c), and (e) show that the 2E3 Vision Encoder outperforms 3D Vision Encoder (+2.3\% BLEU-1), and combining both encoders further improves performance by using complementary hybrid 3D features.

\begin{figure}[t]
\centering  
\includegraphics[width=1.0\linewidth]{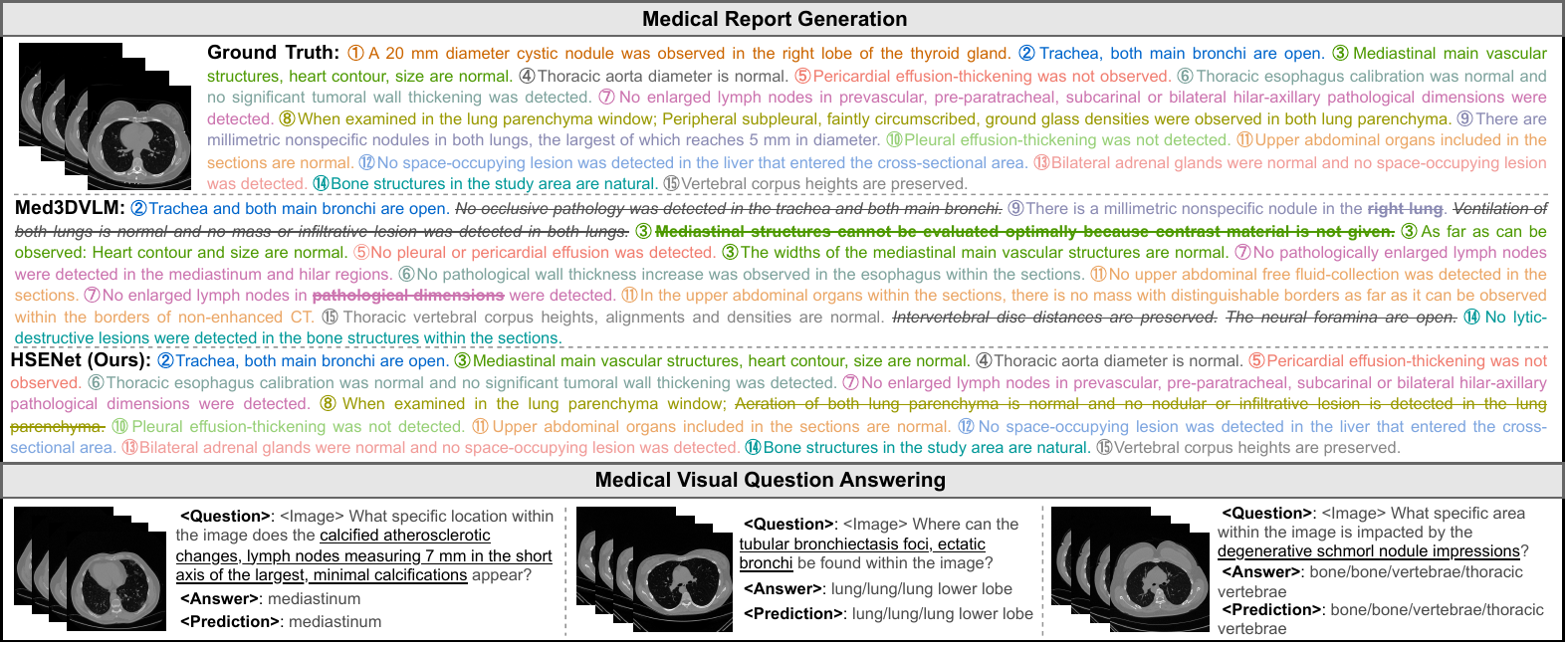}  
\vspace{-10pt}  
\caption{
Visualization of 3D CT report generation and medical VQA. Different colors in reports highlight distinct diagnostic findings. \sout{Strikethrough} marks incorrect predictions, while \textit{italicized words} indicate generated contents absent from ground truth reports.
}
\label{report_visualize}  
\end{figure}

\textbf{Visualization of Report Generation.}
Figure~\ref{report_visualize} visualizes reports generated by our model and the most advanced Med3DVLM. We find that Med3DVLM exhibits notable errors and hallucinations in diagnosing 3D organs, highlighting the challenges of understanding 3D spatial patterns. 
In contrast, our HSENet produces more accurate diagnoses and identifies key structures, such as ``\textit{bilateral adrenal glands}'' and ``\textit{thoracic aorta}'', which Med3DVLM overlooks. These results further demonstrate the strength of our hybrid visual contexts in capturing 3D spatial information.

\subsection{Results on Medical VQA}
We also assess the model’s spatial reasoning capability via medical visual question answering.
As shown in Table~\ref{table:vqa}, our approach surpasses prior methods, with accuracy gains of +3.21\% (major classes) and +1.38\% (minor classes) over Med3DVLM.
Notably, we find that using only the 2E3 Visual Encoder already yields notable gains over the 3D Visual Encoder competitor, achieving 1.47\%$\uparrow$ on BLEU-1 and 1.21\%$\uparrow$ on Major Class Accuracy.
This suggests that our model relies more on the local 3D features given by 2E3 Visual Encoder for reasoning spatial locations. 
By aggregating both the local and global 3D representations, our HSENet captures richer visual contexts and achieves the best performance, which is aligned with the findings of \citet{Huang2021GLoRIA}.
Qualitative results in Figure~\ref{report_visualize} further demonstrate HSENet’s ability to infer precise 3D locations in VQA scenarios.

\section{Conclusion}
We present HSENet, a novel 3D medical vision-language model that bridges the visual perception and projection to understand complex 3D spatial structures for CT diagnosis.
HSENet introduces dual 3D vision encoders to perceive both global volumetric context and local anatomical details, and designs a spatial packer to project 3D spatial features into the LLM’s semantic space via compact, informative tokens.
We conduct comprehensive evaluations on benchmark datasets across 3D multi-modal retrieval, report generation, and medical VQA tasks. HSENet achieves state-of-the-art performance in both visual representation learning and diagnostic text generation.
We believe this work can provide promising insights toward unified 3D image-report understanding and inspire further research in enhancing computer-aided CT diagnosis.

\section*{Acknowledgements}
This work was supported in part by the National Natural Science Foundation of China under Grant 61906007, 62276010, and 62306253, in part by the Guangdong Natural Science Fund-General Programme under Grant 2024A1515010233.









\appendix

\section{Visual Token Compression Sensitivity Study}
To further explore the token compression and 3D spatial preservation capabilities of our spatial packer, we perform sensitivity experiments comparing various strides $(S_d, S_w, S_h)$ to control the number of compressed tokens during the encoding of low-resolution points $\mathbf{V}_{lr}^{G}\in \mathbb{R}^{S_d \times S_w \times S_h \times d_v}$ (See section 3.3 in the manuscript).
$S_d$, $S_w$, and $S_h$ denote the count of partitioned voxels in the spatial dimensions of the volume feature, $\hat{D}$, $\hat{W}$, and $\hat{H}$, with each voxel having dimensions $(\frac{\hat{D}}{S_d}, \frac{\hat{W}}{S_w}, \frac{\hat{H}}{S_h})$.

\begin{table}[h]
\centering
\scriptsize  
\caption{
Performance comparison across different numbers of visual tokens in the spatial packer for report generation.
Ratios (e.g., X\%$\downarrow$ or Y\%$\uparrow$) indicate the degree of token reduction or expansion relative to the default setting.
}
\label{table:stride_parameter}
\setlength{\tabcolsep}{2.9pt}{
\begin{tabular}{ccccccccccc}
\toprule
\textbf{Token Number} &
  \textbf{Stride} &
  \textbf{BLEU-1} &
  \textbf{BLEU-2} &
  \textbf{BLEU-3} &
  \textbf{BLEU-4} &
  \textbf{ROUGE-1} &
  \textbf{ROUGE-L} &
  \textbf{METEOR} &
  \textbf{BERT-Score} &
  \textbf{RaTE-Score}\\ \hline \\[-2.0ex]
32 (75\%$\downarrow$)  & (8,2,2) & 60.95          & 36.37          & 25.33          & 19.77          & 55.96          & 37.78          & 43.92          & 88.69          & 63.16          \\
64 (50\%$\downarrow$)  & (4,4,4) & 61.43          & 36.73          & 25.66          & 20.11          & 55.92          & 37.96          & 43.72          & 88.73          & 63.40          \\
\rowcolor[HTML]{EFEFEF} 
128 (default)\rule{0pt}{2.0ex} & (8,4,4) & 62.89          & \textbf{39.47} & \textbf{29.11} & \textbf{24.01} & \textbf{56.50} & \textbf{40.63} & \textbf{44.75} & \textbf{88.99} & \textbf{64.99} \\
256 (100\%$\uparrow$) & (4,8,8) & \textbf{63.41} & 38.31          & 27.09          & 21.59          & 55.73          & 38.80          & 43.06          & 88.84          & 63.55          \\ [-0.6ex]
\bottomrule
\end{tabular}
}
\end{table}

As shown in Table~\ref{table:stride_parameter}, when the number of visual tokens is compressed to 32, the model performance decreases compared to our original configuration (128 tokens per spatial packer), with BLEU-1 and ROUGE-L dropping by 1.94\% and 2.85\%, respectively. 
Nonetheless, its clinical performance is still equivalent to the variant of spatial pooling-based projector with 128 tokens (RaTE-Score: 63.16\%, see Table 2(b) in our main manuscript), which demonstrates that our spatial packer can preserve the clinical relevance of generated context, even under extreme token compression.
As the visual token count increases from 32 to 128 (32 $\rightarrow$ 64 $\rightarrow$ 128), we observe a gradual improvement in performance, particularly in BLEU-4 (from 19.77\% to 20.11\% to 24.01\%, 4.24\% $\uparrow$ in total). 
This suggests that the model's ability to generate coherent and contextually accurate text improves with more visual tokens, emphasizing the importance of token quantity for text generation quality.
However, increasing the token count beyond 128, particularly to 256, results in degraded performance (1.44\%$\downarrow$ of RaTE-Score).
The reason may be due to the reduced voxel size $(\frac{\hat{D}}{S_d}, \frac{\hat{W}}{S_w}, \frac{\hat{H}}{S_h})$, which impairs the V2P-CA module’s capacity to capture salient high-resolution structures.
Consequently, the spatial features become less discriminative, leading to decreased overall performance.

\section{Evaluations on BIMCV-R Dataset}
We conduct additional experiments on the BIMCV-R dataset~\cite{Chen2024BIMCV}, a benchmark for 3D medical report generation.
This dataset comprises 8,069 3D CT volumes (over 2 million slices), each paired with a corresponding medical report.
Following the preprocessing protocol of~\citet{Lai2024E3D}, we use 6,766 volume-report pairs for training and 752 for testing.
We reuse our dual 3D visual encoders pretrained on the CT-RATE dataset~\cite{Hamamci2024DevelopingGF} without further updates to extract volume features from BIMCV-R.
Only the spatial packer and LoRA layers are fine-tuned for task adaptation.

The results of report generation are presented in Table~\ref{table:BIMCV_caption}.
Notably, despite using frozen visual encoders ($\mathbf{E}_{\mathrm{3d}}(\cdot)$ and $\mathbf{E}_{\mathrm{2e3}}(\cdot)$) pretrained exclusively on the CT-RATE dataset, HSENet achieves the best performance across all evaluation metrics, including a 14.28\% increase in BLEU-1 over E3D-GPT.
This demonstrates the effectiveness of our pretraining strategy in capturing valuable spatial patterns from 3D CT volumes.
It is also interesting to find that E3D-GPT, which adopts self-reconstruction for visual pretraining, obtains the second-best results in BERTScore, ROUGE-1, and METEOR (81.78\%, 23.93\%, and 13.62\%, respectively). 
This suggests that, under limited data conditions (BIMCV-R contains only 14.3\% as many training samples as CT-RATE), self-reconstruction may enable the learning of more expressive medical representations than CLIP-style pretraining, thus benefiting downstream report generation.
These findings point to a promising direction for future research: integrating self-reconstruction with vision-language alignment to further enhance the understanding of 3D medical visual features.

\section{Evaluation of Clinical Efficiency in VQA}
We evaluate the clinical efficiency of HSENet in answering questions across various anatomical locations, including the heart, breast, and lung.
Ten key body locations are selected based on the official categories provided by RadGenome-ChestCT dataset~\cite{Zhang2024RadGenome}.
We compare HSENet against several strong baselines, including M3D-LaMed~\cite{Bai24M3D}, Med-2E3~\cite{Shi24Med2E3}, and Med3DVLM~\cite{Xin2025Med3DVLMAE}, as well as different variants of HSENet using single or dual visual encoders.

\begin{figure}[t]
\centering  
\includegraphics[width=1.0\linewidth]{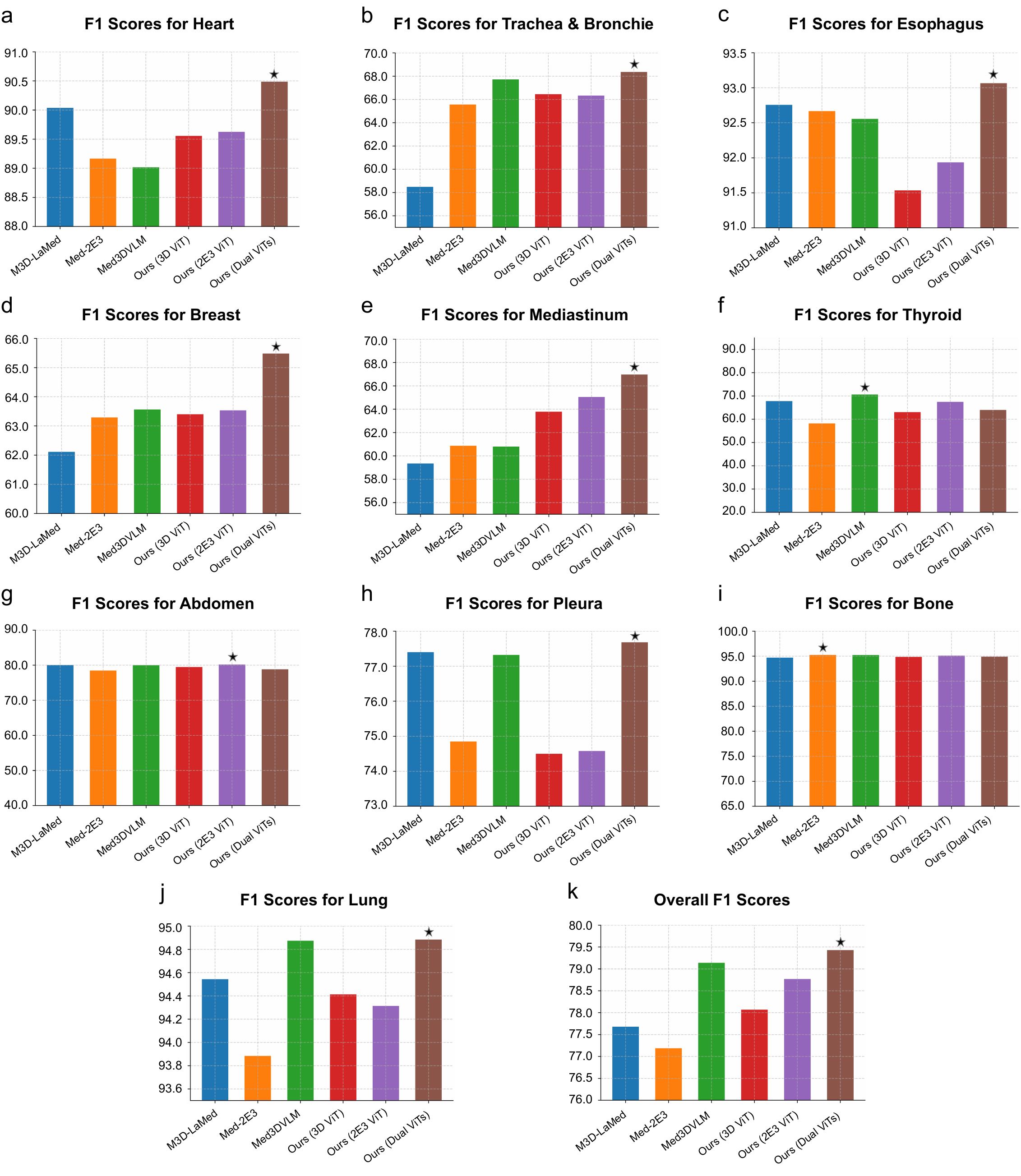}  
\vspace{-10pt}  
\caption{
Experiments on the clinical effectiveness of VQA across different body locations. 
Colored bars represent different methods, with the star symbol ($\star$) indicating the highest F1 score. 
\textit{3D ViT}, \textit{2E3 ViT}, and \textit{Dual ViTs} denote the use of our 3D Visual Encoder $\mathbf{E}_{\mathrm{3d}}(\cdot)$, 2E3 Visual Encoder $\mathbf{E}_{\mathrm{2e3}}(\cdot)$, and both encoders within the proposed HSENet, respectively.
Each subfigure records the F1 score for a specific body location, while the final subfigure (k) shows the average clinical performance across all locations.
}

\label{VQA_F1}  
\end{figure}

As shown in the final subfigure of Figure~\ref{VQA_F1}, HSENet achieves the highest overall F1 score (79.42\%) among all methods, demonstrating its ability to interpret diverse 3D spatial patterns in complex clinical reasoning tasks.
It performs especially promising on anatomically stable regions such as the heart (90.48\%), lung (94.88\%), and breast (65.47\%).
In contrast, we also noticed that the performance on structurally irregular regions like bone is less optimal.
Despite a strong F1 score of 94.82\%, HSENet slightly underperforms Med-2E3 by 0.35\%.
This may be attributed to the uniform voxel partitioning used in our spatial packer, which limits its adaptability to highly variable skeletal structures.
Therefore, introducing adaptive voxel partitioning could offer a promising future direction for enhancing spatial encoding and improving performance in regions with complex anatomical variation.

\begin{table}[t]
\centering
\scriptsize  
\caption{
Experiments on medical report generation on the BIMCV-R dataset~\cite{Chen2024BIMCV}.
\textbf{Bold} indicates the best performance.
† denotes the reproduced models.
}
\label{table:BIMCV_caption}
\setlength{\tabcolsep}{4.0pt}{
\begin{tabular}{lccccccccc}
\toprule
\multicolumn{1}{c}{\textbf{Methods}} &
  \textbf{BLEU-1} &
  \textbf{BLEU-2} &
  \textbf{BLEU-3} &
  \textbf{BLEU-4} &
  \textbf{ROUGE-1} &
  \textbf{ROUGE-L} &
  \textbf{METEOR} &
  \textbf{BERT-Score} &
  \textbf{RaTE-Score} \\ \hline\\[-2.0ex]
RadFM\cite{Wu23RadFM}     & 0.83  & /    & /    & /    & 3.87  & /     & 1.98  & 78.21 & /     \\
CT-CHAT\cite{Hamamci2024DevelopingGF}   & /     & /    & /    & /    & /     & /     & /     & /     & /     \\
M3D-LaMed\cite{Bai24M3D} & 16.43 & /    & /    & /    & 21.44 & /     & 11.38 & 81.63 & /     \\
E3D-GPT\cite{Lai2024E3D}   & 18.19 & /    & /    & /    & 23.93 & /     & 13.62 & 81.78 & /     \\
Med-2E3\cite{Shi24Med2E3}†  & 27.32 & 5.99 & 2.01 & 0.79 & 14.77 & 10.55 & 8.40  & 80.01 & 34.65 \\
Med3DVLM\cite{Xin2025Med3DVLMAE}† & 31.13 & 5.29 & 1.55 & 0.66 & 20.09 & 12.05 & 11.55 & 81.73 & 32.92 \\
\rowcolor[HTML]{EFEFEF} 
HSENet (Ours)\rule{0pt}{2.0ex} &
  \textbf{32.47} &
  \textbf{7.33} &
  \textbf{2.71} &
  \textbf{1.43} &
  \textbf{24.88} &
  \textbf{14.98} &
  \textbf{14.67} &
  \textbf{82.50} &
  \textbf{36.13} \\ [-0.6ex]
\bottomrule
\end{tabular}
}
\end{table}

\begin{figure}[t]
\centering  
\includegraphics[width=1.0\linewidth]{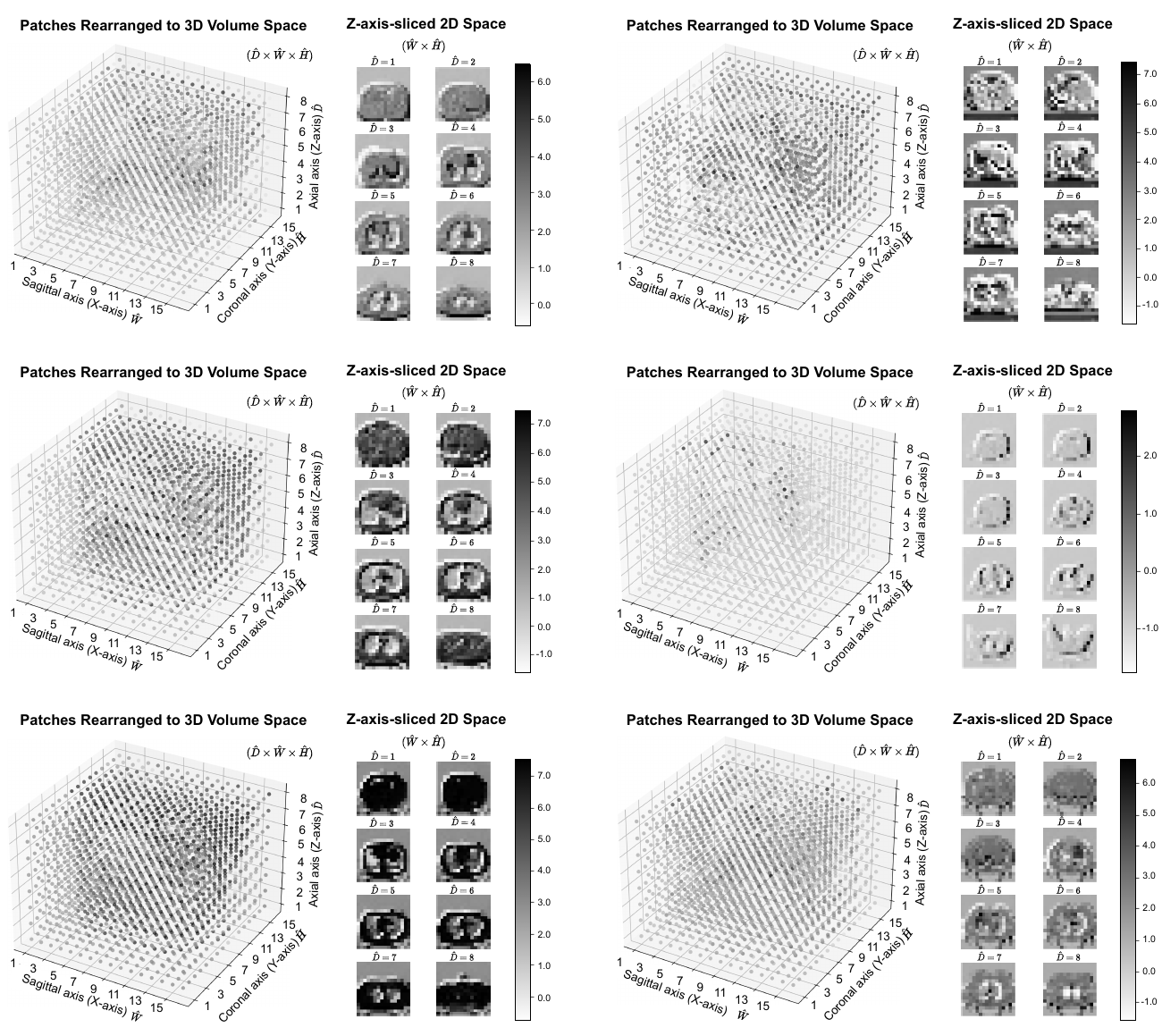}  
\vspace{-10pt}  
\caption{
Additional visualizations of 3D patch scores generated by the 2E3 Visual Encoder $\mathbf{E}_{\mathrm{2e3}}(\cdot)$.
Darker colors indicate higher scores. 
Both 3D views (patches rearranged into the original volume space) and 2D views (axial slices along the Z-axis at different depth levels $\hat{D}$) are provided to illustrate the spatial distribution of scores.
}
\label{More_scores}  
\end{figure}

\begin{figure}[t]
\centering  
\includegraphics[width=1.0\linewidth]{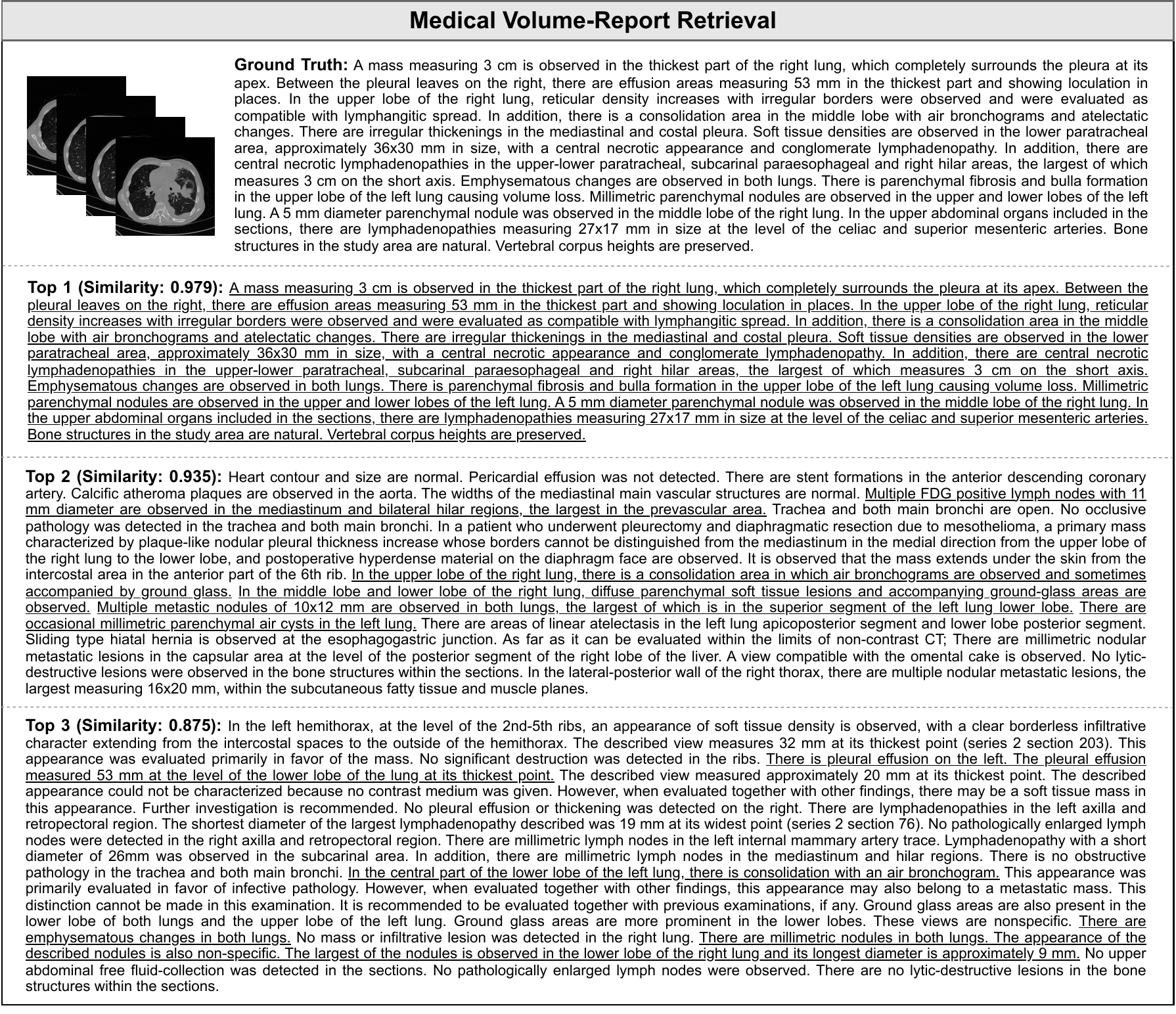}  
\vspace{-10pt}  
\caption{
Visualization of medical volume-to-report retrieval.
The 2E3 visual encoder $\mathbf{E}_{\mathrm{2e3}}(\cdot)$ and the text decoder $\mathbf{E}_{\text{text}}^{s_2}(\cdot)$ is utilized to encode 3D volume and report features, respectively. 
For each input volume, the top-3 retrieved reports are shown to assess retrieval quality. \underline{Underlined} sentences highlight key findings consistent with the ground-truth report.
}
\label{retrieval}  
\end{figure}

\begin{figure}[ht]
\centering  
\includegraphics[width=1.0\linewidth]{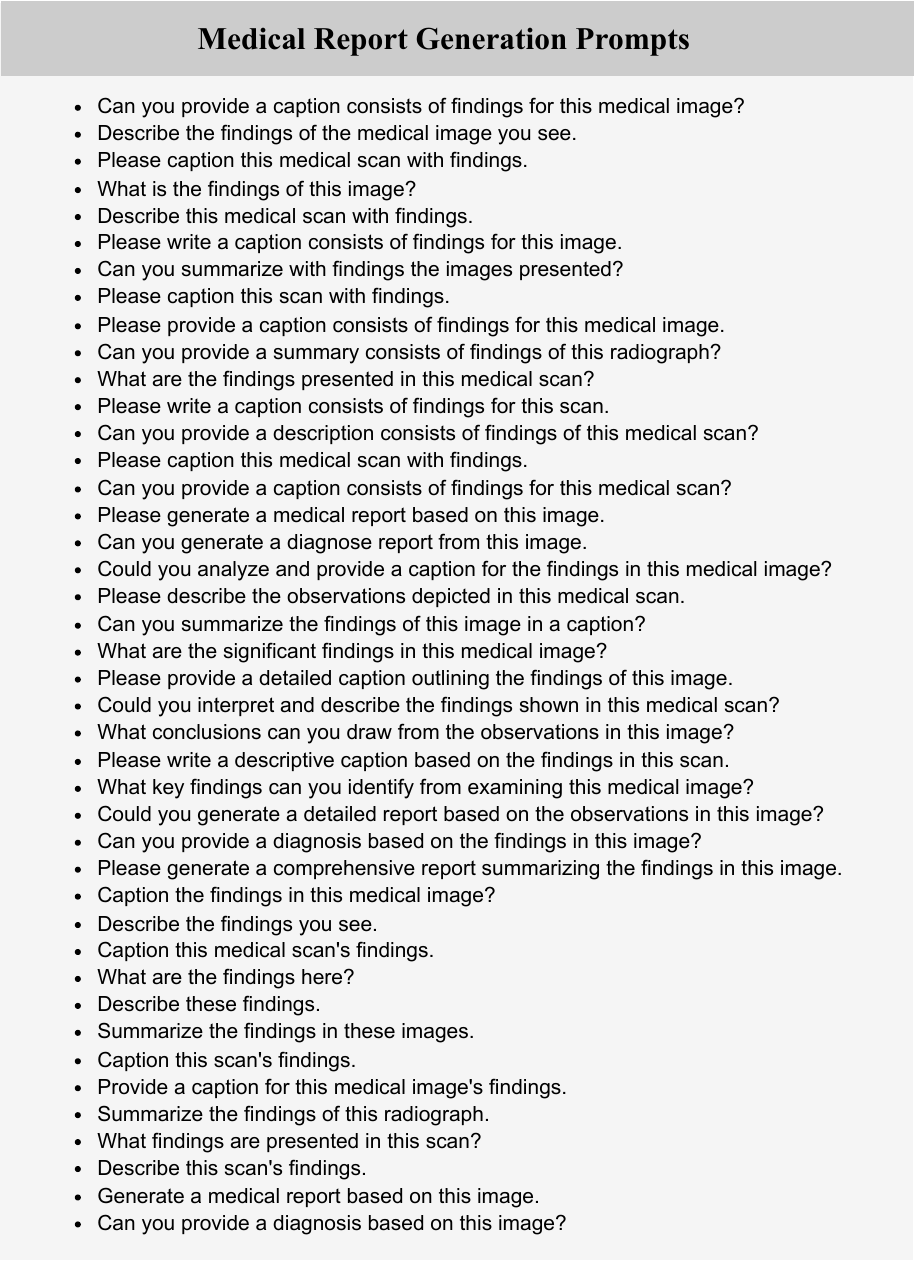}  
\vspace{-10pt}  
\caption{
Textual prompts for medical report generation follow the format of~\citet{Bai24M3D}. To enhance the instruction-following capability of HSENet, prompts are randomly assigned to samples during training.
}
\label{mrg_prompts}  
\end{figure}

\begin{figure}[ht]
\centering  
\includegraphics[width=1.00\linewidth]{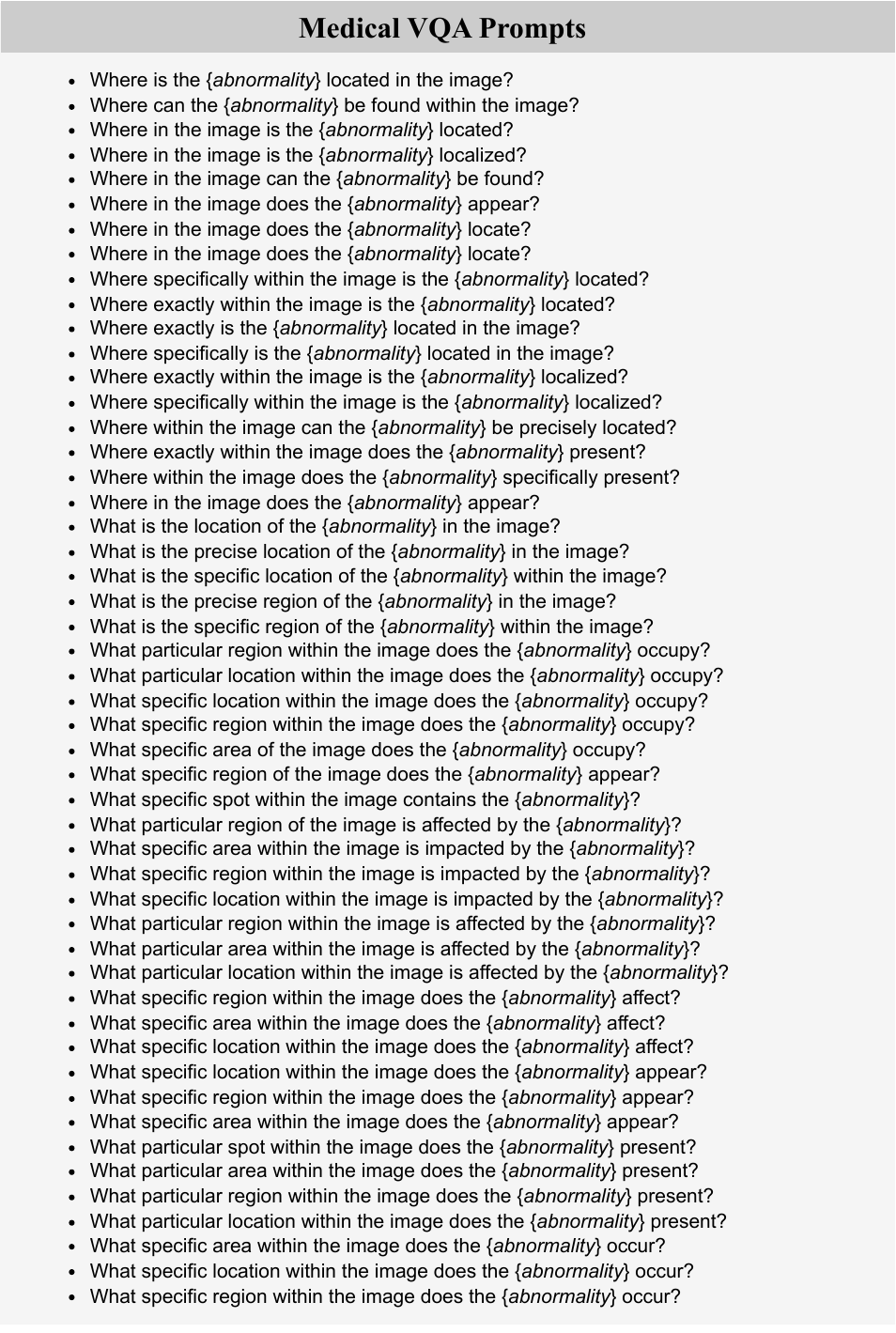}  
\vspace{-10pt}  
\caption{
Textual prompts for medical VQA follow the format of~\citet{Zhang2024RadGenome}. To ensure HSENet’s instruction-following ability, prompts are randomly assigned to training samples. The placeholder \textit{\{abnormality\}} indicates where location-specific abnormalities are inserted.
}
\label{vqa_prompts}  
\end{figure}

\begin{wraptable}{r}{0.5\textwidth}  
\centering
\caption{
Results of inference efficiency analysis.  \textit{s/item} refers to seconds per item.
The inference time for human radiologists is provided from ~\citet{Sexauer2022TimeIM} for reference.
}
\vspace{10pt}
\label{table:efficiency}
\setlength{\tabcolsep}{6.0pt}{
\begin{tabular}{lcc}
\toprule
 & \textbf{Report Generation} & \textbf{VQA} \\ 
 \midrule
Ours         & 3.59 \textit{s/item}                  & 1.11 \textit{s/item}    \\
Radiologist  & $\sim$950.40 \textit{s/item}            & /            \\ \bottomrule
\end{tabular}
}
\end{wraptable}

\section{Training, Inference, and Computational Resources}
This section outlines the detailed configurations and computational resources used for model training and inference.

\textbf{Vision-Language Pretraining.} 
We perform two-stage pretraining on the CT-RATE dataset~\cite{Hamamci2024DevelopingGF}. Both pretraining stages are trained for 50 epochs using 8 NVIDIA RTX 3090 GPUs, with approximately 23 GB of memory use and 24 data loader workers per GPU. Each stage requires roughly 26-28 hours of training.

\textbf{MLLM Fine-tuning.}
We apply 8-bit quantization to the LLM and fine-tune it with LoRA~\cite{hu2021lora}. Fine-tuning is conducted on 8 NVIDIA RTX 3090 GPUs. For the report generation task, we train on the CT-RATE dataset~\cite{Hamamci2024DevelopingGF} for 6 epochs, using 22 GB memory per GPU and 22 workers per GPU, requiring approximately 14 hours. For the VQA task, we train on the RadGenome-ChestCT dataset~\cite{Zhang2024RadGenome} for 4 epochs with similar resource settings, taking about 22 hours in total.

\textbf{Inference Efficiency.}
Table~\ref{table:efficiency} presents the inference latency of HSENet.
Our HSENet generates diagnostic reports in 3.59 seconds per instance and answers VQA queries in 1.11 seconds on average.
Compared to human radiologists, who require approximately 950.40 seconds (15.84 minutes) per report~\citep{Sexauer2022TimeIM}, HSENet achieves a $\sim264\times$ speedup in report generation.

\section{Additional Qualitative Analysis}

\textbf{3D Patch Scoring.}
Figure~\ref{More_scores} shows the scoring distributions of 3D patches generated by our 2E3 Visual Encoder $\mathbf{E}_{\mathrm{2e3}}(\cdot)$.
These distributions exhibit substantial variation across samples, suggesting that our model effectively captures both intra-sample patch differences and inter-sample visual variability.
This adaptive scoring strategy can effectively enhance the discriminability of learned 3D visual representations, thereby boosting the model’s performance in multi-modal retrieval and text generation tasks.

\textbf{Volume-Report Retrieval.}
To evaluate retrieval performance, we provide qualitative examples using the pretrained stage-2 multi-modal encoders, i.e., 2E3 visual encoder $\mathbf{E}_{\mathrm{2e3}}(\cdot)$ and text decoder $\mathbf{E}_{\text{text}}^{s_2}(\cdot)$, to extract features from 3D volumes and medical reports.
As shown in Figure~\ref{retrieval}, given a query volume, our method retrieves its ground-truth report from a test set of 3,039 volume-report pairs with high confidence (0.979). 
Notably, the top-2 and top-3 retrieved reports also demonstrate strong semantic similarity to the ground truth, despite minor lexical variations (e.g., ``There are emphysematous changes in both lungs.'' vs. ``Emphysematous changes are observed in both lungs.'').
These results indicate that our 2D-enhanced 3D learning framework effectively captures cross-modal correlations, enabling accurate and semantically aligned volume-report retrieval.

\section{Textual Prompts for Model Training}
To empower HSENet with instruction-following capabilities towards 3D medical tasks, e.g., medical report generation and medical VQA, we adopt a diverse set of textual prompts during training.
For report generation, we follow the protocol of~\citet{Bai24M3D}, utilizing 42 distinct prompt templates (see Figure~\ref{mrg_prompts}). A prompt is randomly selected for each training instance to improve the robustness and generalization of the vision-language model. These prompts are consistently applied across ablation studies and baseline comparisons to ensure fairness.
For the VQA task, we similarly employ 50 prompt types from~\citet{Zhang2024RadGenome} (see Figure~\ref{vqa_prompts}), enabling HSENet to generalize across a wide range of question formats.

\section{Limitations}
Clinical diagnosis typically relies on a combination of 3D visual data and rich contextual information, including patient history, clinical interviews, and electronic health records.
While this work tackles a core challenge of learning generalizable 3D spatial representations and yields strong performance across a range of downstream tasks, we do not explicitly address the organization or integration of diverse contextual clinical data during pretraining or fine-tuning.
This omission may lead to suboptimal diagnostic text generation in more complex, real-world scenarios, potentially undermining clinical reliability.
Therefore, a key direction for future work is the effective collection and integration of multi-modal, multi-source clinical data to improve the robustness and reliability of 3D diagnostic systems.

\section{Dataset License}
This work uses publicly available benchmark datasets: CT-RATE~\cite{Hamamci2024DevelopingGF} (CC-BY-NC-SA 4.0 License), RadGenome-ChestCT~\cite{Zhang2024RadGenome} (CC-BY 4.0 License), and BIMCV-R~\cite{Chen2024BIMCV} (MIT License).
All licenses permit usage for research purposes.
We fully comply with the respective license terms, and all datasets are used solely for research without any modification or repackaging.

\newpage
\section*{NeurIPS Paper Checklist}

\begin{enumerate}

\item {\bf Claims}
    \item[] Question: Do the main claims made in the abstract and introduction accurately reflect the paper's contributions and scope?
    \item[] Answer: \answerYes{} 
    \item[] Justification: The abstract and introduction state the scope and contributions of the paper, including the introduction of the HSENet, its effective pretraining strategy and spatial packer for visual perception and projection, and the substantial performance gains achieved across diverse downstream tasks.

\item {\bf Limitations}
    \item[] Question: Does the paper discuss the limitations of the work performed by the authors?
    \item[] Answer: \answerYes{} 
    \item[] Justification: The paper includes a section on limitations in supplemental material, outlining the scope of the framework, and areas for future improvement.

\item {\bf Theory assumptions and proofs}
    \item[] Question: For each theoretical result, does the paper provide the full set of assumptions and a complete (and correct) proof?
    \item[] Answer: \answerNA{} 
    \item[] Justification: The paper does not present theoretical results that necessitate formal assumptions or proofs.

    \item {\bf Experimental result reproducibility}
    \item[] Question: Does the paper fully disclose all the information needed to reproduce the main experimental results of the paper to the extent that it affects the main claims and/or conclusions of the paper (regardless of whether the code and data are provided or not)?
    \item[] Answer: \answerYes{} 
    \item[] Justification: The paper provides comprehensive details on datasets, experimental setups, and methodologies used, ensuring that the results can be reproduced accurately.

\item {\bf Open access to data and code}
    \item[] Question: Does the paper provide open access to the data and code, with sufficient instructions to faithfully reproduce the main experimental results, as described in supplemental material?
    \item[] Answer: \answerYes{} 
    \item[] Justification: The code is included in the supplemental material and will be open-sourced upon acceptance to support the 3D medical vision-language understanding research community.

\item {\bf Experimental setting/details}
    \item[] Question: Does the paper specify all the training and test details (e.g., data splits, hyperparameters, how they were chosen, type of optimizer, etc.) necessary to understand the results?
    \item[] Answer: \answerYes{} 
    \item[] Justification: The paper specifies relevant experimental details, including data splits, number of samples, and hyperparameters, ensuring transparency and reproducibility of the results.

\item {\bf Experiment statistical significance}
    \item[] Question: Does the paper report error bars suitably and correctly defined or other appropriate information about the statistical significance of the experiments?
    \item[] Answer: \answerNo{} 
    \item[] Justification: Error bars are not reported due to the high computational cost of 3D medical volume-report modeling.

\item {\bf Experiments compute resources}
    \item[] Question: For each experiment, does the paper provide sufficient information on the computer resources (type of compute workers, memory, time of execution) needed to reproduce the experiments?
    \item[] Answer: \answerYes{} 
    \item[] Justification: Detailed information on computational resources, including workers, memory, and inference time, is provided in the supplementary materials to ensure reproducibility.
    
\item {\bf Code of ethics}
    \item[] Question: Does the research conducted in the paper conform, in every respect, with the NeurIPS Code of Ethics \url{https://neurips.cc/public/EthicsGuidelines}?
    \item[] Answer: \answerYes{} 
    \item[] Justification: This research aligns with the NeurIPS Code of Ethics, ensuring responsible conduct throughout the study.

\item {\bf Broader impacts}
    \item[] Question: Does the paper discuss both potential positive societal impacts and negative societal impacts of the work performed?
    \item[] Answer: \answerYes{} 
    \item[] Justification: The paper discusses potential societal impacts, with positive impacts mentioned in the introduction, and negative impacts mentioned in the limitations section (see supplement materials).
    These include the benefits of medical-assisted models as well as risks related to medical hallucinations in generated responses.
    
\item {\bf Safeguards}
    \item[] Question: Does the paper describe safeguards that have been put in place for responsible release of data or models that have a high risk for misuse (e.g., pretrained language models, image generators, or scraped datasets)?
    \item[] Answer: \answerYes{} 
    \item[] Justification: The paper discusses potential limitations in supplemented materials, such as risks of medical hallucinations and incorrect diagnoses in 3D medical image analysis.

\item {\bf Licenses for existing assets}
    \item[] Question: Are the creators or original owners of assets (e.g., code, data, models), used in the paper, properly credited and are the license and terms of use explicitly mentioned and properly respected?
    \item[] Answer: \answerYes{} 
    \item[] Justification: The paper properly credits the creators of existing assets used and states the licenses and terms of use.

\item {\bf New assets}
    \item[] Question: Are new assets introduced in the paper well documented and is the documentation provided alongside the assets?
    \item[] Answer: \answerNA{} 
    \item[] Justification: The paper does not introduce any new assets.

\item {\bf Crowdsourcing and research with human subjects}
    \item[] Question: For crowdsourcing experiments and research with human subjects, does the paper include the full text of instructions given to participants and screenshots, if applicable, as well as details about compensation (if any)? 
    \item[] Answer: \answerNA{} 
    \item[] Justification: The paper does not involve crowdsourcing or research with human subjects.

\item {\bf Institutional review board (IRB) approvals or equivalent for research with human subjects}
    \item[] Question: Does the paper describe potential risks incurred by study participants, whether such risks were disclosed to the subjects, and whether Institutional Review Board (IRB) approvals (or an equivalent approval/review based on the requirements of your country or institution) were obtained?
    \item[] Answer: \answerNA{} 
    \item[] Justification: The paper does not involve crowdsourcing or research with human subjects.

\item {\bf Declaration of LLM usage}
    \item[] Question: Does the paper describe the usage of LLMs if it is an important, original, or non-standard component of the core methods in this research? Note that if the LLM is used only for writing, editing, or formatting purposes and does not impact the core methodology, scientific rigorousness, or originality of the research, declaration is not required.
    \item[] Answer: \answerYes{} 
    \item[] Justification: The LLM is utilized in the HSENet as the language decoder, which is not the core innovation of this research. This is clarified in the method and experiment sections.

\end{enumerate}

\end{document}